\documentclass{article}

\usepackage{amsmath}
\usepackage{algorithm}
\usepackage{algpseudocode}
\usepackage[utf8]{inputenc}
\usepackage{graphicx}

\usepackage{graphicx} 

\usepackage{hyperref}
\usepackage{amsmath}
\usepackage{amssymb}
\usepackage{multirow}

\usepackage{tikz}
\usetikzlibrary{matrix}

\usepackage[utf8]{inputenc}
\usepackage{scrextend}

\usepackage{graphicx}
\usepackage{subcaption}

\usepackage{colortbl}
\usepackage{pifont}
\usepackage{array}
\usepackage{multirow}
\usepackage{xcolor}

\usepackage{booktabs}

\usepackage{algorithm}
\usepackage{algpseudocode}

\usepackage{mdframed}

\usepackage{empheq}
\usepackage{fancybox}

\usepackage{hyperref}

\usepackage{pgfplots}
\pgfplotsset{compat=1.16} 

\usepackage[title]{appendix}
\usepackage{etoolbox}

\def\email#1{{\texttt{#1}}}
\providecommand{\keywords}[1]
{
  \small	
  \textbf{\textit{Keywords---}} #1
}

\title{R-ParVI: Particle-based variational inference through lens of rewards}
\author{Yongchao Huang \footnote{Author email: \email{yongchao.huang@abdn.ac.uk}}}
\date{July 2024}

\begin{document}

\maketitle

\begin{abstract}
    A reward-guided, gradient-free ParVI method, \textit{R-ParVI}, is proposed for sampling partially known densities (e.g. up to a constant). R-ParVI formulates the sampling problem as particle flow driven by rewards: particles are drawn from a prior distribution, navigate through parameter space with movements determined by a reward mechanism blending assessments from the target density, with the steady state particle configuration approximating the target geometry. Particle-environment interactions are simulated by stochastic perturbations and the reward mechanism, which drive particles towards high density regions while maintaining diversity (e.g. preventing from collapsing into clusters). R-ParVI offers fast, flexible, scalable and stochastic sampling and inference for a class of probabilistic models such as those encountered in Bayesian inference and generative modelling. 
\end{abstract}


\section{Introduction}

In many fields, such as statistics, machine learning, and computational science, a fundamental challenge is to sample from complex, intractable probability densities, such as Bayesian posteriors, which are often known only up to an uncomputable normalizing constant. These distributions play a central role in applications like Bayesian inference, uncertainty quantification, and generative modeling, where understanding the underlying parameter space is essential. Modern datasets and models increasingly exhibit high dimensionality, multimodality, and intricate geometric structures. However, traditional sampling techniques, such as Markov Chain Monte Carlo (MCMC), face substantial limitations: their slow convergence and high computational cost in high-dimensional or poorly mixing scenarios render them impractical for many real-world problems. Alternatively, variational inference (VI) provides a faster approximation but often sacrifices accuracy by relying on simpler, parametric distributions that fail to fully capture the target’s complexity, especially in multimodal cases.

To overcome these challenges, we propose \textit{R-ParVI}, a novel particle-based variational inference (ParVI) method that integrates a reward-guided framework inspired by reinforcement learning (RL). R-ParVI reimagines sampling as an optimization process, where a population of particles evolves under a composite reward function that balances the target density with an entropy-driven diversity term. This approach enables R-ParVI to efficiently guide particles towards high-probability regions while preventing collapse into overly concentrated clusters, ensuring a faithful representation of the target distribution’s structure. Unlike many existing methods (e.g. SVGD \cite{Liu2016SVGD}), R-ParVI is gradient-free, facilitating its applicability to scenarios where gradient information is unavailable or computationally expensive. Additionally, its stochastic and parallelizable nature enhances scalability, making it well-suited for large-scale probabilistic models. This work, in its current version, presents the R-ParVI methodology conceptually; experimental evaluation of its effectiveness and efficiency on some benchmark tasks and real-world applications shall follow in later updated works \footnote{Preliminary experiments in Aug. 2024 confirm the efficiency and benchmark accuracy of the RL-inspired ParVI methods.}.

\section{Related work}

\paragraph{Sampling methods} \footnote{This list is by no means complete.} Sampling from complex probability distributions, especially those that are partially known or unnormalized, is a fundamental challenge in statistical inference and probabilistic modeling. Classical sampling techniques have laid the groundwork for this field. For instance, MCMC methods, such as Metropolis-Hastings (MH) \cite{Metropolis1953}, Gibbs sampling \cite{Geman1984}, Langevin Monte Carlo (LMC) \cite{Roberts1996}, and Hamiltonian Monte Carlo (HMC) \cite{Duane1987}, generate samples by constructing Markov chains that converge to the target distribution over time. These methods are versatile, with gradient-free approaches like MH and slice sampling \cite{Neal2003} being particularly useful for unnormalized densities encountered in Bayesian inference, while gradient-based methods like HMC and LMC leverage derivative information for improved efficiency, albeit at a higher computational cost. However, MCMC methods often face limitations such as slow convergence and poor scalability in high-dimensional spaces. Other traditional techniques, such as inverse transform sampling \cite{Devroye1986}, rejection sampling \cite{vonNeumann1951}, and importance sampling \cite{Kahn1949}, offer solutions when the full \textit{probability density function} (pdf) is known, but they struggle with intractable distributions.

To address these challenges, \textit{particle-based variational inference} (ParVI) methods have emerged as a promising class of techniques that approximate target distributions using a set of particles \footnote{Particle systems can be interactive (e.g. SVGD \cite{Liu2016SVGD}, EParVI \cite{huang2024EParVI}, SPH-ParVI \cite{huang2024SPHParVI}, MPM-ParVI \cite{huang2024MPMParVI}) or non-interactive as in this work.}. Methods like Stein variational gradient descent (SVGD \cite{Liu2016SVGD}) evolve particles by minimizing the Kullback-Leibler (KL) divergence to the target distribution using gradient information, offering scalability and efficiency in high-dimensional settings. Sequential Monte Carlo (SMC \cite{Doucet2001}) and particle-based energetic VI (EVI) \cite{Wang2021EVI} similarly use particle ensembles, reducing the curse of dimensionality that concerns traditional MCMC. More recently, physics-inspired ParVI methods have gained attention, for example, electrostatics-based ParVI (EParVI \cite{huang2024EParVI}), which employs electrostatic principles, and smoothed particle hydrodynamics (SPH-ParVI \cite{huang2024SPHParVI}) and material point method (MPM-ParVI \cite{huang2024MPMParVI}) based ParVI methods, which adapt concepts from fluid or solid mechanics to guide particle movements. These methods vary in their reliance on gradients: SVGD and EVI are gradient-based, while adaptations like EParVI explore gradient-free dynamics. R-ParVI builds on these ParVI foundations by introducing a novel reward-driven mechanism inspired by reinforcement learning, aiming to enhance both the flexibility and effectiveness of sampling from complex distributions.

\paragraph{Reinforcement learning and sampling} Sampling methods have been employed in \textit{reinforcement learning} (RL) to estimate state-action distributions from data, bridging probabilistic inference with decision-making. For example, MCMC techniques have been adapted to sample from posterior distributions over policies or value functions, while SVGD has been used to approximate distributions over states and actions, leveraging its gradient-based particle dynamics to refine RL outcomes \cite{Chung2020,Depthfirstlearning2019}. \textit{Reward design} plays a critical role in RL, as it provides the feedback mechanism that guides an agent towards optimal behavior. Well-crafted reward functions balance exploration and exploitation, shaping the learning process to achieve desired goals. Research in this area has explored hand-crafted rewards, inverse RL for reward inference, and shaping techniques to stabilize training \cite{Devidze2021,Ng2000,Ma2024}.

While sampling techniques like MCMC and SVGD have proven useful in RL, the reverse, i.e. using RL principles to enhance sampling, has been under-explored until recently \cite{Chung2020}. Traditional ParVI methods, such as SVGD and EParVI, focus on minimizing statistical discrepancies (e.g. KL divergence or \textit{MMD} metric) or simulating physical interactions, but they lack an adaptive mechanism to balance density-seeking (converging to high-probability regions) and diversity maintenance (exploring the distribution broadly). R-ParVI addresses this gap by incorporating RL-inspired reward mechanisms into the ParVI framework. By treating particle movements as actions guided by a reward signal, R-ParVI leverages the intuitive structure \footnote{Formulating the sampling problem fully as a RL problem will be followed in a separate work; here we just leverage the reward mechanism, avoiding defining \textit{states} and \textit{actions}.} of RL to dynamically adjust sampling behavior, offering a hybrid approach that combines the scalability of ParVI with the adaptability of RL. This methodological innovation distinguishes R-ParVI from prior work and positions it as a novel tool for sampling in scenarios where gradient information is limited or computational resources are constrained.

\section{R-ParVI: methodology}

We have the following problem: \textit{given a $d$-dimensional density $p(\mathbf{x})$, which can be complex, intractable (e.g. a Bayesian posterior), or unnormalized, we want to infer its geometry by generating quasi-samples via variational routines}. This task is inherently challenging because $p(\mathbf{x})$ may be of multimodality, high dimensionality, or computational intractability, making traditional sampling methods like MCMC inefficient or impractical. To overcome these difficulties, we can use particle-based variational inference (ParVI) techniques which suffer less from the curse of dimensionality while offering flexibility and scalability. 

To simulate the evolution of a particulate system, the R-ParVI algorithm introduces the concept of reward in reinforcement learning (RL) to guide particle movements towards approximating a target distribution. The reward function is designed as:

\begin{equation} \label{eq:reward}
    R(\mathbf{x}_p) = \alpha \cdot \tilde{p}(\mathbf{x}_p) + \beta \cdot \left( -\tilde{p}(\mathbf{x}_p) \log(\tilde{p}(\mathbf{x}_p)) \right),
\end{equation}
which comprises two key components: a \textit{density} term proportional to the target density $\tilde{p}(\mathbf{x})$, driving particles towards high-probability regions, and a \textit{diversity} term which promotes particle spread by penalizing excessive clustering through an entropy-based penalty. The weights $\alpha$ (e.g. 0.6) and $\beta$ (e.g. 0.4) balance the trade-off between density-seeking and diversity maintenance.

The reward $R(\mathbf{x}_p)$ is then used to guide updates of velocity and position for each particle. At iteration $t$, the algorithm evaluates potential moves for a particle $p$ by introducing a small perturbation $\boldsymbol{\delta}_p$, leading to a test position $\mathbf{x}_p' = \mathbf{x}_p^{(t-1)} + \boldsymbol{\delta}_p$. The reward at this test position $\mathbf{x}_p'$ is compared to the reward at current position $\mathbf{x}_p^{(t-1)}$: if $R(\mathbf{x}_p') > R(\mathbf{x}_p^{(t-1)})$, the particle’s velocity is updated to encourage movement in the direction of the perturbation:
  
\begin{equation}
  \mathbf{v}_p^{(t)} = \mathbf{v}_p^{(t-1)} + \eta \cdot \boldsymbol{\delta}_p,
\end{equation}
where $\eta$ is the learning rate controlling the step size of the velocity adjustment. Otherwise, if the reward does not improve, the velocity is damped to slow down the particle’s movement:
  
\begin{equation}
  \mathbf{v}_p^{(t)} = \gamma \cdot \mathbf{v}_p^{(t-1)},
\end{equation}
with $\gamma$ (e.g. 0.9) as the damping (discount) factor that reduces the velocity magnitude.

Following the velocity update, the particle’s position is then updated using the new velocity, augmented by a stochastic exploration term $\mathbf{e}_p \sim \mathcal{N}(0, \epsilon^2 \mathbf{I}_d)$, where $\epsilon$ controls the scale of exploration and $\mathbf{I}_d$ is the identity matrix in $d$-dimensional space:

\begin{equation}
    \mathbf{x}_p^{(t)} = \mathbf{x}_p^{(t-1)} + \mathbf{v}_p^{(t)} + \mathbf{e}_p
\end{equation}

To ensure the particle remains within the parameter space, the new position may be clipped to predefined bounds, e.g. $\mathbf{x}_p^{(t)} \in [-L, L]^d$. This iterative process, i.e. evaluating test positions, adjusting velocities based on reward comparisons, and updating positions with exploration, enables the particles to collectively approximate the target distribution $\tilde{p}(\mathbf{x})$ while maintaining diversity, making R-ParVI a flexible and efficient framework for variational inference. The full R-ParVI algorithm is presented in Algo.\ref{algo:RParVI}.

\begin{algorithm}[H]
\fontsize{8}{8}
\small
\caption{R-ParVI: Reward-based ParVI}
\label{algo:RParVI}
\begin{itemize}
\item{\textbf{Inputs}: a target (possibly unnormalised) density $\tilde{p}(\mathbf{x})$, total number of particles $M$, particle dimension $d$, initial proposal distribution $p^0(\mathbf{x})$, total number of iterations $T$, learning rate $\eta$, exploration rate $\epsilon$, velocity discount rate $\gamma$.}
\item{\textbf{Outputs}: Particles located at positions $\{\mathbf{x}_p \in \mathbb{R}^d\}_{p=1}^M$ whose empirical distribution approximates the target density $\tilde{p}(\mathbf{x})$, along with reward history for convergence analysis.}
\end{itemize}
\vskip 0.01in
1. \textbf{\textit{Initialize}} particles. [$\mathcal{O}(Md)$]
    \begin{itemize}
        \item Initialize $M$ particles $\{\mathbf{x}_p^{(0)}\}_{p=1}^M$ by sampling from an initial proposal distribution $\mathbf{x}_p \sim p^0(\mathbf{x})$. e.g. $p^0(\mathbf{x}) = \mathcal{U}(-L, L)^d$ with bound $L$.
        \item Initialize particle velocities $\{\mathbf{v}_p^{(0)}\}_{p=1}^M$ to zero, i.e., $\mathbf{v}_p^{(0)} = \mathbf{0} \in \mathbb{R}^d$.
    \end{itemize}

2. \textbf{\textit{Update}} particle positions. \\
For each iteration $t = 1, 2, \dots, T$, repeat: \\
        (1) Compute the reward $R(\mathbf{x}_p^{(t-1)})$ for each particle $p$: [$\mathcal{O}(Md)$]
        \[
        R(\mathbf{x}_p^{(t-1)}) = \alpha \cdot \tilde{p}(\mathbf{x}_p^{(t-1)}) + \beta \cdot \left( -\tilde{p}(\mathbf{x}_p^{(t-1)}) \log(\tilde{p}(\mathbf{x}_p^{(t-1)})) \right)
        \]
        where $\tilde{p}(\mathbf{x}_p^{(t-1)})$ is the (unnormalised) target density evaluated at particle $\mathbf{x}_p^{(t-1)}$, $\alpha$ and $\beta = 1 - \alpha$ are weights balancing the density and diversity terms (e.g. $\alpha = 0.6$, $\beta = 0.4$). \\
        (2) For each particle $p$: [$\mathcal{O}(Md)$]
            \begin{itemize}
                \item generate a displacement exploration term $\mathbf{e}_p \sim \mathcal{N}(\mathbf{0}, \epsilon^2 \mathbf{I}_d)$, where $\epsilon$ is the exploration rate.
                \item generate a displacement perturbation $\boldsymbol{\delta}_p \sim \mathcal{N}(\mathbf{0}, 0.1^2 \mathbf{I}_d)$.
                \item evaluate the test position $\mathbf{x}_p' = \mathbf{x}_p^{(t-1)} + \boldsymbol{\delta}_p$. 
                \item compute the test reward $R(\mathbf{x}_p')$. 
                \item update velocity:
                \[
                \mathbf{v}_p^{(t)} =
                \begin{cases}
                    \mathbf{v}_p^{(t-1)} + \eta \cdot \boldsymbol{\delta}_p & \text{if } R(\mathbf{x}_p') > R(\mathbf{x}_p^{(t-1)}), \\
                    \gamma \cdot \mathbf{v}_p^{(t-1)} & \text{otherwise}.
                \end{cases}
                \]
                $\gamma$ is a velocity damping (discount) factor, e.g. $\gamma=0.9$.
                \item update the particle position:
                \[
                \mathbf{x}_p^{(t)} = \mathbf{x}_p^{(t-1)} + \mathbf{v}_p^{(t)} + \mathbf{e}_p,
                \]
                clip $\mathbf{x}_p^{(t)}$ to the range $[-L, L]^d$ to maintain bounded exploration, where $L$ is a user-defined bound.  
            \end{itemize}
        (3) Compute and store the mean reward:
        \[
        \bar{R}^{(t)} = \frac{1}{M} \sum_{p=1}^M R(\mathbf{x}_p^{(t)}).
        \]
        Record particle trajectories and reward histories.

3. \textbf{\textit{Return}} the final particle positions $\{\mathbf{x}_p^{(T)}\}_{p=1}^M$, their empirical distribution (via histogram or KDE), and the reward history $\{\bar{R}^{(t)}\}_{t=0}^{T-1}$ for convergence analysis.
\end{algorithm}

\paragraph{Computational efficiency} Designed for scalability and efficiency, R-ParVI operates with $M$ particles (e.g. 100) in $d$-dimensional space, achieving a computational complexity \footnote{It totals $\mathcal{O}(M T d)$ for $M$ particles over $T$ iterations with $d$ being the dimensionality of the problem. Note that, this complexity estimate doesn't take into account the cost of evaluating the data-involved term such as likelihood.} of $\mathcal{O}(Md)$ \textit{per iteration}, mainly dominated by individual reward evaluations. Unlike some ParVI methods requiring costly $\mathcal{O}(M^2 d)$ distance matrix computations \footnote{If distance-based diversity metrics are used, the computational cost will increase to $\mathcal{O}(M^2d)$.}. (e.g. SVGD evaluates kernalised, pairwise distances), R-ParVI uses a local entropy term for diversity, avoiding pairwise interactions \footnote{Ignoring particle-particle interactions can also be an disadvantage in securing accuracy.}. Key hyperparameters, such as the displacement exploration rate $\epsilon$ (e.g. 0.1), velocity learning rate $\eta$ (e.g. 0.1) and simulation steps $T$ (e.g. 1000), allow fine-tuning of convergence and exploration. 

R-ParVI is gradient-free, relying on (relative) reward comparisons to guide particle movement, making it suitable for sampling partially known densities. As absolute and exact rewards are not required, the R-ParVI algorithm can be used to sample from unnormalised densities. Convergence can be diagnosed via the \textit{stepwise reward curve} (similar to the epoch-wise cumulative reward curve in RL). It is also flexible, efficient and therefore scalable. R-ParVI can be parallelized \footnote{ParVI methods are, in general, convenient for parallelism.} effectively due to the \textit{independence of particle updates}: a particle updates its positions using velocity whose value is based on the reward which is independent of other particles \footnote{This is also a drawback of the R-ParVI algorithm, as it lacks of particle-particle interactions.} - reward only relies on the local 'environment', i.e. landscape of the target density. In step 2.1, the reward $R(\mathbf{x}_p^{(t-1)})$ depends solely on the particle’s own position $\mathbf{x}_i^t$ and the target density $\tilde{p}(\mathbf{x}_i^t)$. Since $\tilde{p}(\mathbf{x}_i^t)$ is a fixed function evaluated independently for each particle, this step can be computed simultaneously for all $M$ particles. In step 2.2, generating $\boldsymbol{\delta}_p$, computing the test position $\mathbf{x}_p'$, evaluating $R(\mathbf{x}_p')$, and updating the velocity $\mathbf{v}_p^{(t)}$ are all operations that rely only on the individual particle’s state (i.e. its position $\mathbf{x}_p^{(t-1)}$, velocity $\mathbf{v}_p^{(t-1)}$, and the perturbation $\boldsymbol{\delta}_p$). There are no dependencies between particles, making these steps fully parallelisable. The exploration term $\mathbf{e}_p$ is sampled independently for each particle, and the position update $\mathbf{x}_p^{(t)} = \mathbf{x}_p^{(t-1)} + \mathbf{v}_p^{(t)} + \mathbf{e}_p$ followed by clipping is a local operation. Again, no inter-particle interactions exist. In step 2.3, computing the aggregated, mean reward requires collecting $R(\mathbf{x}_p^{(t)})$ from all particles. While this step involves combining results, it can be efficiently parallelized using a \textit{reduction operation} (e.g. summing rewards across threads and dividing by $M$), a standard technique in parallel computing. 

R-ParVI isn’t strictly RL though. In RL, agent(s) learns to make decisions by interacting with an environment or with other agents over time. Agent observes current state, selects an action based on a (learned or prescribed) policy, receives a reward, and aims to maximize the cumulative reward through trial and error. One needs to define \textit{state space} (the set of all possible states the environment can be in), \textit{action space} (the set of all possible actions the agent can take), \textit{policy} (a strategy mapping states to actions, e.g. a policy network), the \textit{reward function} (feedback from the environment based on the agent’s actions), \textit{value function} (an estimate of the expected cumulative reward from a state or state-action pair), and an \textit{initial state}. RL often employs techniques like experience replay (storing and reusing past experiences) and decision-making algorithms (e.g. REINFORCE, DDPG, PPO, SAC, Q-learning, etc) to stabilize and optimize learning. These elements are not defined in R-ParVI. One may also notice its connection to the MH algorithm. If $\beta=\gamma=0$ and $\mathbf{e}_p=\mathbf{0}$, R-ParVI is similar to \textit{multiple-chain MH} with symmetric transition kernel. 

\paragraph{Limitations} R-ParVI doesn't account for particle-particle interactions; it only takes into account particle-environment interactions via stochastic perturbations and the rewarding mechanism. Ignoring particle-particle interactions can be an disadvantage in securing accuracy. Also, more informed moves, can be designed to accelerate convergence, e.g. encoding gradient and neighbourhood information into $\boldsymbol{\delta}_p$ and $\boldsymbol{e}_p$.

\section{Conclusion and future work}
We introduce R-ParVI, a particle-based variational inference method, which employs a reward-driven approach to iteratively guide particles toward high-density regions of a target distribution $p(\mathbf{x})$ while incorporating an entropy-based term to ensure diversity and prevent mode collapse. This gradient-free method, relying on reward directions rather than gradient information, balances density-seeking with exploration, and offers a flexible, scalable and computationally efficient framework for sampling complex, intractable distributions. It also produces quasi-samples that effectively capture the geometric structure of the target distribution. 
Current design assumes independence between particles; future enhancements could accounting for inter-particle correlations to further refine the sampling process. Also, formulating sampling fully as a RL problem can potentially enhance its adaptability and capability to handle increasingly complex probabilistic models.

\bibliography{reference}

\end{document}